# The meaning-frequency law in Zipfian optimization models of communication

*Ramon Ferrer-i-Cancho*[1]

**Abstract.** According to Zipf's meaning-frequency law, words that are more frequent tend to have more meanings. Here it is shown that a linear dependency between the frequency of a form and its number of meanings is found in a family of models of Zipf's law for word frequencies. This is evidence for a weak version of the meaning-frequency law. Interestingly, that weak law (a) is not an inevitable of property of the assumptions of the family and (b) is found at least in the narrow regime where those models exhibit Zipf's law for word frequencies.

## 1. Introduction

The relationship between the frequency of a word and its number of meanings follows Zipf's law of meaning distribution: words that are more frequent tend to have more meanings (Zipf 1945; Baayen & Moscoso del Prado Martín 2005; Ilgen & Karaoglan 2007; Crossley et al. 2010; Hernández-Fernández et al. 2016). In his pioneering research, Zipf defined two laws where $\mu$, the number of meaning of a word, is the response variable (Zipf 1945; Zipf 1949). One law where the predictor variable is $i$, the rank of a word (the most frequent word has rank 1, the 2nd most frequent word has rank 2 and so on), i.e.

$$\mu \propto i^{-\gamma}, \qquad (1)$$

where $\gamma \approx 1/2$. Another law where the predictor variable is $f$, the frequency of a word,

$$\mu \propto f^{\delta}, \qquad (2)$$

where $\delta$ is a constant satisfying $\delta \approx 1/2$. Zipf (1949) referred to Eq. 1 as the law of meaning distribution in his most famous book while he referred to Eq. 2 as the meaning-frequency relationship in a less popular article (Zipf, 1945). Eq. 1 and Eq. 2 describe, through different predictor variables, the qualitative tendency of the number of meanings of a word to increase as its frequency increases (assuming $\gamma > 0$ and $\delta > 0$).

Zipf (1945) derived Eq. 2 from Eq. 1 and Zipf's law for word frequencies (with rank as the predictor variable), i.e.

$$f \propto i^{-\alpha}, \qquad (3)$$

---
[1] Complexity and Quantitative Linguistics Lab. LARCA Research Group. Departament de Ciències de la Computació, Universitat Politècnica de Catalunya (UPC). Campus Nord, Edifici Omega, Jordi Girona Salgado 1-3. 08034 Barcelona, Catalonia (Spain). Phone: +34 934134028. E-mail: rferrericancho@cs.upc.edu.

where $\alpha \approx 1$ (Zipf 1945; Zipf 1949). Zipf's derivation of Eq. 2 is revisited in the Appendix. After Zipf's untimely decease, some researchers investigated Eq. 2 and provided support for it independently from Zipf's law for word frequencies. i.e. Eq. 2 (Baayen & Moscoso del Prado Martín 2005; Crossley et al. 2010; Hernández-Fernández et al. 2016) while others provided further empirical support for Eq. 1 (Ilgen & Karaoglan 2007).

$\mu$ is a measure of the polysemy of a word. If the mapping of words into meanings is regarded as a bipartite graph joining word vertices with meaning vertices (Ferrer-i-Cancho et al. 2005), $\mu$ defines the (semantic) degree of a word in that network.

The target of the preset article is the meaning-frequency law. Eq. 2 with $\delta \approx 1/2$ defines a strong version of the meaning-frequency law. A weak version of the meaning-frequency law can be defined simply as a positive correlation between $\mu$ and $f$. Notice that we are not assuming a Pearson correlation here, which is a measure of linear association (Conover 1999). Instead, we have in mind an association measure that can capture non-linear dependencies, e.g., Spearman rank correlation (Conover 1999; Zhou et al. 2003). The definition and use of a weak version of the law is justified for various reasons. First, looks of a pure power law of the form of Eq. 2 can be deceiving. This is a lesson of research on Menzerath-Altmann law in genomes (Ferrer-i-Cancho et al. 2013a), Heaps' law in texts (Font-Clos & Corral 2015) or the degree distribution of protein interaction and metabolic networks (Stumpf & Ingram 2005; Stumpf et al. 2005). Admittedly, Zipf's laws of meaning (Eq. 1 and 2) are not among the most investigated statistical laws of language and a mathematical argument illuminating the origins of both laws is not forthcoming. Thus, the basis for their current formulation is purely empirical. Second, the weak meaning-frequency law allows one to remain neutral about the actual dependency between $\mu$ and $f$. This neutral formulation has been adopted for research on Menzerath-Altmann's law in genomes (Ferrer-i-Cancho et al. 2013a) and Zipf's law of abbreviation in animal behavior (Ferrer-i-Cancho et al. 2013b and references therein). Third, the weak version allows for a unified approach to human language and the communicative behavior of other species. A positive correlation between frequency and behavioral context (a proxy for meaning) has been found in dolphin whistles (Ferrer-i-Cancho & McCowan 2009). Fourth, the weak version provides enough flexibility to allow for parsimonious models of language, models that reproduce more than one law of language at least qualitatively (Ferrer-i-Cancho, 2013).

In spite of the generality of the weak version of the law, its formulation imposes some hidden constraints. First, the fact that the law is an empirical law, imposes that only words that have non-zero probability matter for deciding if a theoretical model agrees with the law. Words that have zero probability are not observable. This second constraint might seem somewhat farfetched but Zipfian optimization models of communication can generate words with zero probability (Ferrer-i-Cancho & Díaz-Guilera 2007; Dickman et al. 2012; Prokopenko et al. 2010; Ferrer-i-Cancho 2013). Second, the definition of a proper correlation between $\mu$ and $f$ (e.g., Pearson correlation, Spearman rank correlation) needs at least two different values of $\mu$ and at least two different values of $f$; otherwise, the variance of $\mu$ and that of $f$ are undefined. To see it recall that the Pearson correlation between two variables $X$ and $Y$ (the ranks of $\mu$ and the ranks of $f$ in case of the Spearman rank correlation between $X$ and $Y$) is defined as

$$r = \frac{COV(X,Y)}{\sigma(X), \sigma(Y)}, \qquad (4)$$

where $COV(X,Y)$ is the covariance of $X$ and $Y$ and $\sigma(X)$ and $\sigma(Y)$ are the standard deviation of $X$ and $Y$, respectively. If $\sigma(X) = 0$ or $\sigma(Y) = 0$ the correlation is undefined.

If one wishes to determine if a theoretical model (e.g., Ferrer-i-Cancho & Solé 2003; Ferrer-i-Cancho 2005) agrees with the weak version of the law, frequencies must be replaced by probabilities. In that case, the definition of a proper correlation needs that there at least two different word probabilities. For the reasons explained above, words that have zero probability are excluded. Thus, the weak law cannot be defined properly in a communication system where only one word has non-zero probability or all words are equally likely. The weak law cannot be defined either in a system where only one word has non-zero degree or all words have the same degree.

Notice that the constraint of non-zero variance in the values of $\mu$ and $f$ also concerns the strong meaning-frequency law (if $f$ does not vary, Eq. 2 is not the only possibility for the relationship between $\mu$ and $f$).

Here we investigate if a family of Zipfian optimization models of communication (Ferrer-i-Cancho & Solé 2003; Ferrer-i-Cancho 2005) is able to reproduce some version of the meaning-frequency law. The family was conceived to investigate Zipf's law for word frequencies. It will be shown that those models yield Eq. 2 with $\delta = 1$ thus satisfying only the weak meaning-frequency relationship.

## 2. The family of Zipfian optimization models

The family of models departs from the assumption that there is a repertoire of $V_S$ forms, $s_1,...,s_i,...,s_{V_S}$ and a repertoire of $V_R$ meanings, $r_1,...,r_i,...,r_{V_R}$ that are associated through a binary matrix $A = \{a_{ij}\}$: $a_{ij} = 1$ if the $s_i$ and $r_j$ are associated ($a_{ij} = 0$ otherwise). The models of that family share also the assumption that the probability that a form $s_i$ is employed to refer to meaning $r_j$ is

$$p(s_i | r_j) = \frac{a_{ij}}{\omega_j}, \quad (5)$$

where

$$\omega_j = \sum_{i=1}^{V_S} a_{ij} \quad (6)$$

is the degree of the $j$-th meaning. The convention that $p(s_i|r_j) = 0$ when $\omega_j = 0$ is adopted. By definition, the marginal probability of $s_i$ is

$$p(s_i) = \sum_{j=1}^{V_R} p(s_i, r_j) = \sum_{j=1}^{V_R} p(s_i | r_j) p(r_j). \quad (7)$$

The models of that family diverge by making further assumptions about $p(r_j)$. While one model (model B) assumes that $p(r_j)$ is given, e.g., $p(r_j)=1/V_R$ assuming that no meaning is disconnected (Ferrer-i-Cancho & Solé 2003), another model (model A) assumes that (Ferrer-i-Cancho 2005)

$$p(r_j) = \frac{\omega_j}{M} \quad (8)$$

being $M$ the total number of associations (links)

$$M = \sum_{j=1}^{V_R} \omega_j. \qquad (9)$$

The Model A/B terminology is borrowed from Ferrer-i-Cancho & Díaz-Guilera (2007). Although the original Model B can easily be extended to allow for disconnected meanings, hereafter Model B with a ban for disconnected meanings is assumed for simplicity.

Applying the assumption of Eq. 5 and the convention on $p(s_i|r_j)$ above, Eq. 7 becomes

$$p(s_i) = \sum_{\substack{j=1 \\ \omega_j > 0}}^{V_R} \frac{a_{ij}}{\omega_j} p(r_j). \qquad (10)$$

For model A, Eq. 8 and 10 lead to

$$p(s_i) = \sum_{\substack{j=1 \\ \omega_j > 0}}^{V_R} \frac{a_{ij}}{\omega_j} \frac{\omega_j}{M} = \frac{1}{M} \sum_{j=1}^{V_R} a_{ij} = \frac{\mu_i}{M}, \qquad (11)$$

where

$$\mu_i = \sum_{j=1}^{V_R} a_{ij} \qquad (12)$$

is the degree of the *i*-th form. Eq. 11 indicates that form probability is proportional to form degree, i.e. model A satisfies Eq. 2 with $\delta = 1$. For model B, Eq. 10 and the assumption that $p(r_j) = 1/V_R$ (recall that no meaning can be disconnected) leads to

$$p(s_i) = \frac{1}{V_R} \sum_{j=1}^{V_R} \frac{a_{ij}}{\omega_j}. \qquad (13)$$

The relationship between the probability of the *i*-th form and degree is not straightforward but it is possible to satisfy a weak version of the meaning-frequency law. Let us impose the following constraint on meaning degrees: $\omega_j = k$ with $0 < k \leq V_R$. This constraint transforms Eq. 13 into

$$p(s_i) = \frac{1}{kV_R} \sum_{j=1}^{V_R} a_{ij} = \frac{1}{kV_R} \mu_i. \qquad (14)$$

Thus, the assumption of identical non-zero meaning degrees produces proportionality between the probability of the *i*-th form and its degree in Model B. Next section will show the utility of the case $k = 1$.

## 3. The weak meaning-frequency law is not inevitable

It has been shown that form probability is proportional to form degree directly in Model A and making further assumptions in Model B but this does not imply that those models are reproducing a weak meaning-frequency law. Some configurations of the matrix A where the weak law is missing will be shown next.

$H(S)$ is defined as the entropy of forms ($S$) and $I(S,R)$ is defined as the mutual information between forms ($S$) and meanings ($R$). The reader is referred to Ferrer-i-Cancho & Díaz-Guilera (2007) for definitions of those information theoretic measures.

If $H(S)$ is minimized, it is well known that then only one form has non-zero probability and non-zero degree (Ferrer-i-Cancho & Díaz-Guilera 2007). Then the variance of form probabilities is zero (recall that form probabilities that are zero are irrelevant for the weak version of the meaning-frequency law) and thus the correlation between form probabilities and semantic degree is not defined (recall Eq. 4). The same problem happens if $I(S,R)$ is maximized. Then the optimal solutions are those where all forms that have non-zero probability have the same degree or the same probability (Ferrer-i-Cancho 2013; Ferrer-i-Cancho & Díaz-Guilera 2007) and thus their variance is zero again.

## 4. A weak meaning-frequency law is possible

### 4.1. Possible in globally optimal configurations

The meaning-frequency law is possible (at least) in the global minima of $H(S|R)$, the conditional entropy of forms when meanings are given. The minima of $H(S|R)$ are characterized by $\omega_j \in \{0,1\}$ for model A (Ferrer-i-Cancho 2013; Ferrer-i-Cancho & Díaz-Guilera 2007) and $\omega_j = 1$ for model B (Ferrer-i-Cancho & Díaz-Guilera 2007; Prokopenko et al. 2010; Dickman et al. 2012). Those minima allow for an arbitrary number of words with non-zero probability/degree (Ferrer-i-Cancho & Díaz-Guilera 2007; Trosso 2008; Prokopenko et al. 2010; Dickman et al. 2012), a requirement of both the strong and weak meaning-frequency law.

The family of models assumes that languages minimize a linear combination of $H(S)$ and $I(S,R)$, i.e.

$$\Omega(\lambda) = -\lambda I(S,R) + (1-\lambda)H(S) \qquad (15)$$

with $0 \leq \lambda \leq 1$.

Eq. 15 is equivalent to (Ferrer-i-Cancho 2005; Ferrer-i-Cancho & Díaz-Guilera 2007)

$$\Omega(\lambda) = (1-2\lambda)H(S) + \lambda H(S|R). \qquad (16)$$

It is not surprising that the mapping of forms into meanings exhibits the principle contrast, the tendency of different forms to contrast in meaning (Clark 1987): the global minima of both $H(S)$ and $H(S|R)$ in Model A and B share $\omega_j \leq 1$ when $V_S \leq V_R$ (Ferrer-i-Cancho & Díaz-Guilera 2007; Ferrer-i-Cancho 2013).

The global minima of $\Omega(\lambda)$ split the range of variation of $\lambda$ into three domains (Ferrer-i-Cancho & Díaz-Guilera 2007):
- $0 \leq \lambda < 1/2$ where only $H(S)$ is minimized. The weak meaning-frequency law is impossible (Section 3).
- $\lambda = 1/2$ where only $H(S|R)$ is minimized. The weak meaning-frequency law is possible (this section).
- $1/2 < \lambda \leq 1$ where only $I(S,R)$ is maximized. The weak meaning-frequency law is impossible (Section 3).

## 4.2. Possible in suboptimal configuration

The appearance of the weak meaning-frequency law is easier when the global minima are not reached. Indeed, those models generate a distribution of forms resembling Zipf's law for word frequencies when $\lambda$ equals $\lambda^*$, a critical value of $\lambda$ when $\Omega(\lambda)$ is optimized by means of an evolutionary algorithm based on a Monte Carlo method at zero temperature (Ferrer-i-Cancho & Solé 2003; Ferrer-i-Cancho 2005; Prokopenko et al. 2010). $\lambda^*$ is typically a value below 1/2 but close to 1/2, i.e.

$$\lambda = 1/2 - \varepsilon \qquad (17)$$

being $\varepsilon$ a small positive quantity, e.g., $\varepsilon = 0.1$ (Ferrer-i-Cancho 2005; Ferrer-i-Cancho & Solé 2003; Prokopenko et al 2010). When $\lambda = \lambda^*$, there is enough variability in form probabilities and their degree to reproduce the meaning-frequency law (Ferrer-i-Cancho & Solé 2003; Ferrer-i-Cancho 2005; Prokopenko et al 2010), a requirement of both the strong and weak meaning-frequency law.

## 4.3. Where a weak meaning-frequency law is found

For model A, a weak meaning-frequency law in the minima of $H(S|R)$ (equivalent to $\Omega(1/2)$) or the suboptimal configurations appearing for $\lambda = \lambda^*$ is not only a possibility but a fact thanks to Eq. 11. For model B, some further reasoning is needed to turn a possibility into a fact. The global minima of $H(S|R)$, i.e. $\omega_j = 1$ with $M>0$, imply $k = 1$ in Eq. 14, which gives

$$p(s_i) = \frac{1}{V_R}\mu_i. \qquad (18)$$

Being the probability of the *i*-th form proportional to its degree, a weak meaning-frequency law is expected in general in the minima of $H(S|R)$ for model B due to the variability of form degrees of these minima: configurations where (a) all forms have the same degree or (b) only one form is connected are unlikely (Trosso 2008; Prokopenko et al. 2010; Dickman et al. 2012). A weak law is also expected in the suboptimal configurations that are obtained for $\lambda = \lambda^*$. This implies that the minimization of $\Omega(\lambda)$ in Eq. 16 is being dominated by the minimization of $H(S|R)$: while the weight of $H(S)$ is small, i.e.

$$(1-2)\lambda^* = 1-2(1/2-\varepsilon) = \varepsilon, \qquad (19)$$

the weight of $H(S|R)$ is relatively large, i.e.

$$\lambda = \lambda^* = 1/2 - \varepsilon, \qquad (20)$$

thanks to Eq. 17. The fact that $H(S|R)$ is much stronger than $H(S)$ is critical for the emergence of the weak meaning-frequency law. The point is that the minimization of $H(S)$ implies the minimization of $H(S|R)$. On the one hand, this is positive for the emergence of the weak law because their minimization promotes in both cases $\omega_j \in \{0,1\}$. On the other hand, this is negative for the emergence of the weak law because we have shown that the minima of $H(S)$ turn the weak meaning-frequency law impossible (Section 3) and the fact that $H(S|R) \leq H(S)$, implies that, if $H(S)$ is minimum, i.e. $H(S) = 0$, then $H(S|R)$ is also minimum, i.e. $H(S|R) = 0$ (Ferrer-i-Cancho & Díaz-Guilera 2007). Being $H(S|R)$ much stronger than $H(S)$ a weak law is expected. Additional support for the arguments comes from the presence of Zipf's law for word frequencies for $\lambda = \lambda^*$ (Ferrer-i-Cancho, 2005; Ferrer-i-Cancho & Solé, 2003). If the minimization of $H(S)$ was dominating, instead of a distribution of this kind one would find one form (or a few forms) taking all probability. This is not what happens (Ferrer-i-Cancho 2005; Ferrer-i-Cancho & Solé 2003; Propopenko et al. 2010).

It is important to note that an inverse-factorial distribution has been derived for $\lambda = \lambda^*$ in Model B (Prokopento et al. 2010) and that this distribution differs from the traditional power-law that is typically used to approximate Zipf's law (Eq. 3). The inverse factorial should be considered as a candidate for in empirical research on Zipf's law (e.g., Li et al. 2010; Font-Clos et al 2013; Gerlach & Altmann 2013).

## 5. Discussion

We have shown some conditions where a weak meaning-frequency law, i.e. Eq. 2 with $\delta = 1$, appears in a family of Zipfian optimization models although the law it is not an inevitable property of the probabilistic definitions. Interestingly, that weak meaning-frequency law emerges (at least) in the narrow range where the models are argued to exhibit Zipf's law for word frequencies. Tentatively, those findings do not imply that a weak meaning-frequency law emerges only under very special circumstances. Suppose that $m=V_S V_R$. The binary association matrix A allows one to produce $2^m$ different mappings of words into meanings. The proportion of mappings (configurations of A) where a Spearman rank correlation is defined and has a positive sign could be large. That should be the subject of future research.

Randomness may facilitate the emergence of the weak law. For instance, consider configurations of the matrix A where the weak law is not found because the correlation is undefined (e.g., those where $H(S)$ is minimum or $I(S,R)$ is maximum). Producing a few random mutations in those configurations, it might be possible to obtain a variance of non-zero probabilities or non-zero degrees that is greater than zero and thus the correlation is defined (recall Eq. 3). Although the correlation is defined, the variation in form probability or form degree may be still too small with regard to real language.

There are many models of Zipf's law for word frequencies (Piantadosi 2014) but as far as we know the family of models reviewed here is the only that illuminates the origin of synchronic properties of language such as the principle of contrast and also dynamic properties such as, the tendency of children to attach new words to unlinked meanings (Ferrer-i-Cancho 2013). It is tempting to conclude that the prediction of a linear relationship between number of meanings and frequency instead of the actual power law dependency of Eq. 2 is a reason to abandon this kind of models (i.e. the current family or variants stemming from it). Although the disagreement between the models examined so far and reality is a serious limitation (and thus we encourage future research), we cannot miss an important point: modern model selection is based on a compromise between parsimony and quality of fit (Burnham & Anderson 2002). To our knowledge, generative models for Eq. 2 are not forthcoming and the predictions of current models of Zipf's law beyond word frequencies are unknown, unexplored or simply impossible (Piantadosi 2014). There is at least one exception: the family of optimization models reviewed here, which is able to shed light on various statistical patterns qualitatively but in one shot from minimal assumptions.

The virtue of that family is not only its parsimonious approach to various laws but also its capacity to unify synchrony (patterns of language such as Zipf's law for word frequencies, a weak meaning-frequency law, the principle of contrast) with diachrony/ontogeny (through the vocabulary learning bias above). The science of the future must be unifying (Morin 1990). Theoretical linguistics cannot be an exception (Alday 2015; Ferrer-i-Cancho 2015).

## Appendix

In his seminal work, Zipf derived Eq. 2 from Eq. 1 (law of meaning distribution) with $\gamma = 1/2$ and Eq. 2 (Zipf's law for word frequencies) with $\alpha = 1$. Here a more general and detailed derivation of Eq. 2 is provided. Notice that Eqs. 1 and 3 give, respectively,

$$i \propto \mu^{-\frac{1}{\gamma}} \tag{21}$$

and

$$i \propto f^{-\frac{1}{\alpha}}. \tag{22}$$

Combining Eqs. 21 and 22 it is obtained

$$\mu \propto f^{\delta} \tag{23}$$

with

$$\delta = \frac{\gamma}{\alpha}. \tag{24}$$

Therefore, $\gamma = 1/2$ and $\alpha = 1$ predict $\delta = 1/2$ as shown originally by Zipf (1945). Interestingly, Eq. 24 shows that Ilgen & Karaoglan's (2007) assumption, namely that $\gamma = \delta$, is only valid if $\alpha = 1$.

**ACKNOWLEDGEMENTS**

We are grateful to A. Corral for helpful comments on the manuscript. We also thank P. Alday and S. Piantadosi for helpful discussions. This work was supported by the grant BASMATI (TIN2011-27479-C04-03) from the Spanish Ministry of Science and Innovation, the grant APCOM (TIN2014-57226-P) from MINECO (Ministerio de Economía y Competitividad) and the grant 2014SGR 890 (MACDA) from AGAUR (Generalitat de Catalunya).